\documentclass{article}

\usepackage{microtype}
\usepackage{graphicx}
\usepackage{booktabs}
\usepackage{hyperref}
\usepackage{amsmath,amssymb,amsthm}
\usepackage{xcolor}

\usepackage[preprint]{icml2026}
\usepackage[capitalize,noabbrev]{cleveref}
\usepackage[most]{tcolorbox}
\definecolor{codepink}{HTML}{E6194B}
\definecolor{codegray}{HTML}{F2F2F2}
\definecolor{txtblue}{HTML}{5BC0F8}
\definecolor{pillfill}{HTML}{DCE8FA}
\definecolor{pillline}{HTML}{6C8EBF}
\definecolor{cmtgray}{HTML}{7A8088}
\definecolor{boxbg}{HTML}{F7F8FA}
\usepackage{listings}
\usepackage[varqu]{zi4}
\tcbuselibrary{listings}
\definecolor{clkw}{HTML}{8A3FFC}
\definecolor{clfn}{HTML}{1F6FEB}
\definecolor{clvar}{HTML}{B45309}
\definecolor{clcmt}{HTML}{8B9097}
\definecolor{clnum}{HTML}{6B7280}
\definecolor{clink}{HTML}{24292F}
\lstdefinestyle{claudelight}{language=Python, basicstyle=\ttfamily\footnotesize\color{clink},
  keywordstyle=\color{clkw}, morekeywords={None}, commentstyle=\color{clcmt},
  emph={f,relu,cross_entropy,arange,sum}, emphstyle=\color{clfn},
  emph={[2]o_now,o_later}, emphstyle={[2]\color{clvar}},
  numbers=left, numberstyle=\ttfamily\footnotesize\color{clnum}, numbersep=9pt, xleftmargin=14pt,
  columns=fullflexible, keepspaces=true, showstringspaces=false, upquote=true, aboveskip=0pt, belowskip=0pt}
\newtcblisting{sensecode}{enhanced, listing only, listing options={style=claudelight},
  colback=white, colframe=black!12, boxrule=0.6pt, arc=6pt, left=2pt, right=6pt, top=5pt, bottom=5pt,
  before skip=8pt, after skip=8pt}
\theoremstyle{plain}
\newtheorem{proposition}{Proposition}
\newcommand{\relu}{\mathrm{ReLU}}
\newcommand{\Sense}{\mathrm{Sense}}
\newcommand{\dto}[2]{d(#1{\to}#2)}
\newcommand{\PLANSCORE}{the round-trip distance $\dto{\hat z}{z_g}+\dto{z_g}{\hat z}$}

\newcommand{\INTRORESULT}{On four pixel environments, Sense identifies irreversible transitions with an AUC between $0.95$ and $1.00$ in every environment and plans successfully in $99$--$100\%$ of trials (\cref{tab:main}).
Under the same encoder, predictor, data and evaluation, LeWorldModel identifies them with an AUC between $0.39$ and $0.59$, close to chance, and each of the three baselines fails in at least one environment, either in danger AUC or in planning.
On TwoRoom, the navigation benchmark of LeWorldModel, Sense reaches the goal in $100\%$ of episodes and LeWorldModel in $91\%$.
In greedy navigation around a cliff, Sense reaches $87\%$ of the goals without a single fall, while LeWorldModel reaches none and falls in $64\%$ of runs (\cref{sec:cliff}).}
\newcommand{\MAINTABLE}{%
\setlength{\tabcolsep}{4pt}
\begin{tabular}{lccccccccc}
\toprule
 & \multicolumn{2}{c}{Cup} & \multicolumn{2}{c}{Cliff} & \multicolumn{2}{c}{Sokoban} & \multicolumn{2}{c}{Pendulum} & TwoRoom \\
\cmidrule(lr){2-3}\cmidrule(lr){4-5}\cmidrule(lr){6-7}\cmidrule(lr){8-9}\cmidrule(lr){10-10}
Method & AUC & Plan & AUC & Plan & AUC & Plan & AUC & Plan & Plan \\
\midrule
EMA target & 0.92{\scriptsize$\pm$0.03} & 0.98{\scriptsize$\pm$0.01} & 0.92{\scriptsize$\pm$0.01} & 0.11{\scriptsize$\pm$0.03} & 0.97{\scriptsize$\pm$0.01} & 0.38{\scriptsize$\pm$0.16} & 0.05{\scriptsize$\pm$0.01} & 0.82{\scriptsize$\pm$0.05} & -- \\
VICReg & 0.72{\scriptsize$\pm$0.04} & 0.98{\scriptsize$\pm$0.01} & 0.63{\scriptsize$\pm$0.12} & \textbf{1.00}{\scriptsize$\pm$0.00} & 0.92{\scriptsize$\pm$0.05} & \textbf{1.00}{\scriptsize$\pm$0.00} & 0.42{\scriptsize$\pm$0.07} & \textbf{1.00}{\scriptsize$\pm$0.00} & -- \\
LeWorldModel & 0.58{\scriptsize$\pm$0.04} & 0.96{\scriptsize$\pm$0.01} & 0.45{\scriptsize$\pm$0.03} & \textbf{1.00}{\scriptsize$\pm$0.00} & 0.59{\scriptsize$\pm$0.10} & \textbf{1.00}{\scriptsize$\pm$0.00} & 0.39{\scriptsize$\pm$0.06} & \textbf{1.00}{\scriptsize$\pm$0.00} & 0.91{\scriptsize$\pm$0.02} \\
\textbf{Sense} & \textbf{1.00}{\scriptsize$\pm$0.00} & \textbf{0.99}{\scriptsize$\pm$0.01} & \textbf{1.00}{\scriptsize$\pm$0.00} & \textbf{1.00}{\scriptsize$\pm$0.00} & \textbf{1.00}{\scriptsize$\pm$0.00} & \textbf{1.00}{\scriptsize$\pm$0.00} & \textbf{0.95}{\scriptsize$\pm$0.02} & \textbf{1.00}{\scriptsize$\pm$0.00} & \textbf{1.00}{\scriptsize$\pm$0.00} \\
\bottomrule
\end{tabular}}
\newcommand{\MAINTEXT}{%
Sense is the only method with a high danger AUC in all four environments with irreversible events, and its planning success matches the best baseline in each (\cref{tab:main}).
On TwoRoom it reaches the goal in every episode with every evaluation seed, while LeWorldModel reaches it in $90$--$94\%$.

LeWorldModel and VICReg spread the embedding evenly, with an effective rank of $25$--$32$ out of $32$.
They plan well, because a three-step search with a good predictor reaches nearby goals in almost any embedding that has not collapsed.
Their embeddings, however, do not reliably separate irreversible steps from ordinary ones, and the danger AUC of LeWorldModel is close to chance.

The target network has the opposite problem.
Its embedding is close to collapse, with an effective rank of $1.6$ on Pendulum, where its danger AUC drops to $0.05$, and on Cliff and Sokoban it recognizes danger but plans successfully in only $11\%$ and $38\%$ of trials.

Sense has an effective rank of $6$--$18$, fewer dimensions than VICReg and LeWorldModel use.
Irreversible steps have the largest return distances in the data, although the loss never refers to irreversibility.}
\newcommand{\CLIFFTABLE}{%
\begin{tabular}{lcc}
\toprule
Method & Reached & Fell \\
\midrule
EMA target & 0.02{\scriptsize$\pm$0.01} & 0.48{\scriptsize$\pm$0.21} \\
VICReg & 0.00{\scriptsize$\pm$0.00} & 0.32{\scriptsize$\pm$0.11} \\
LeWorldModel & 0.00{\scriptsize$\pm$0.00} & 0.64{\scriptsize$\pm$0.10} \\
\textbf{Sense} & \textbf{0.87}{\scriptsize$\pm$0.02} & \textbf{0.00}{\scriptsize$\pm$0.00} \\
\bottomrule
\end{tabular}}
\newcommand{\CLIFFTEXT}{%
With LeWorldModel the greedy agent reaches none of these goals and falls in $64\%$ of runs (\cref{tab:cliff}).
With VICReg it also reaches none and falls in $32\%$, and with a target network it reaches $2\%$ and falls in $48\%$.
The LeWorldModel embedding places the cell beyond the edge close to the cell before it, so the shortest way in the embedding leads over the cliff.
With Sense the agent reaches $87\%$ of the goals and never falls.
It receives no label for the cliff, yet the cells beyond the edge lie far from the goal, because no trajectory ever returns from them.}
\newcommand{\SYMTABLE}{%
\setlength{\tabcolsep}{4pt}
\begin{tabular}{lcccc}
\toprule
 & \multicolumn{2}{c}{Danger AUC} & \multicolumn{2}{c}{Irreversible} \\
\cmidrule(lr){2-3}\cmidrule(lr){4-5}
 & Symmetric & Sense & Symmetric & Sense \\
\midrule
Cup & 0.79{\scriptsize$\pm$0.01} & \textbf{1.00}{\scriptsize$\pm$0.00} & 33{\scriptsize$\pm$5}\% & \textbf{0}{\scriptsize$\pm$0}\% \\
Cliff & 0.86{\scriptsize$\pm$0.05} & \textbf{1.00}{\scriptsize$\pm$0.00} & 7{\scriptsize$\pm$1}\% & \textbf{0}{\scriptsize$\pm$0}\% \\
Sokoban & 0.81{\scriptsize$\pm$0.02} & \textbf{1.00}{\scriptsize$\pm$0.00} & 1{\scriptsize$\pm$0}\% & \textbf{0}{\scriptsize$\pm$0}\% \\
Pendulum & 0.43{\scriptsize$\pm$0.04} & \textbf{0.95}{\scriptsize$\pm$0.02} & \textbf{20}{\scriptsize$\pm$3}\% & 22{\scriptsize$\pm$2}\% \\
\bottomrule
\end{tabular}}
\newcommand{\SYMTEXT}{%
The symmetric variant does not collapse, so the future sampling and the cross-entropy alone already give a useful embedding (\cref{tab:sym}).
The direction identifies irreversible transitions.
With a symmetric distance the return distance equals the forward distance, and the danger AUC falls to $0.79$--$0.86$, and to $0.43$ in Pendulum.
In planning toward the farthest goals, the symmetric variant ends $33\%$ of Cup trials and $7\%$ of Cliff trials with an irreversible event, while Sense ends none.
In Pendulum both end about a fifth of the trials with a fall, $20\%$ and $22\%$, and the difference made by the direction shows in the danger AUC, $0.43$ against $0.95$.}

\icmltitlerunning{The Sense Paradigm for World Models}

\begin{document}

\twocolumn[
\icmltitle{The Sense Paradigm for World Models}

\begin{icmlauthorlist}
\icmlauthor{Haidong Hu}{ujn}
\end{icmlauthorlist}
\icmlaffiliation{ujn}{University of Jinan, Jinan, China}
\icmlcorrespondingauthor{Haidong Hu}{haidonghu.ml@gmail.com}
\icmlkeywords{world models, representation learning, representation collapse, irreversibility, planning}

\vskip 0.3in
]
\printAffiliationsAndNotice{}

\begin{abstract}
Latent world models predict the future in an embedding space rather than in pixels.
An encoder trained only to predict collapses and maps every observation to the same point, so existing models add a mechanism against collapse, such as a target network, a variance penalty or a distribution penalty.
These mechanisms keep the embedding from collapsing, but they leave its distances without meaning, and the model cannot tell a reversible transition from an irreversible one.
We propose Sense, a single cross-entropy loss over the directed distance $\dto{x}{y}=\sum_i \relu(y_i-x_i)$ that replaces these mechanisms.
The loss avoids collapse on its own, and since the encoder no longer needs the predictor, the predictor is fitted afterwards to fixed embeddings by plain regression.
On four pixel environments, the return distance learned by Sense detects irreversible transitions with an AUC between $0.95$ and $1.00$, where LeWorldModel stays between $0.39$ and $0.59$.
On TwoRoom, the navigation benchmark of LeWorldModel, planning with Sense reaches the goal in $100\%$ of episodes against $91\%$ for LeWorldModel.
As the goal moves farther away, the planning failure rate of every baseline grows, while that of Sense stays near zero in Cup, Cliff and Sokoban and is the lowest of all methods in Pendulum.
\end{abstract}

\section{Introduction}

\begin{figure}[t]
\centering
\includegraphics[width=\columnwidth]{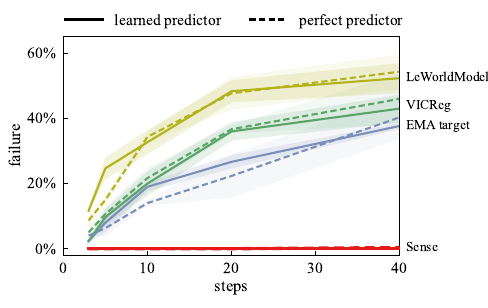}
\caption{Planning failure rate in Cup. Steps is the number of steps from the start to the goal. Solid lines show the failure rate with the learned predictor. Dashed lines show the failure rate when the prediction is exactly correct under the true physics. Lines are the mean over three seeds and shaded bands span one standard deviation.}
\label{fig:cup}
\end{figure}

\begin{figure*}[t]
\centering
\includegraphics[width=\textwidth]{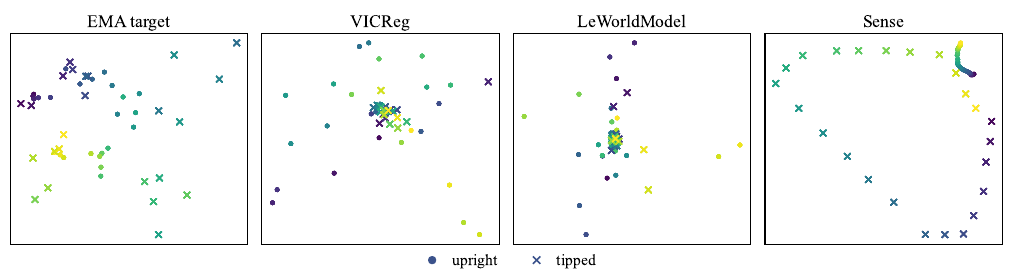}
\caption{\textbf{Same encoder, same predictor, same data. Only the representation objective changes.} Two-dimensional PCA projections of the embeddings of a cup at 200 positions, upright ($\circ$) and tipped ($\times$), coloured by position. Each panel is scaled to fit its box. An EMA target, VICReg and LeWorldModel avoid collapse but mix tipped cups with upright ones. With Sense, upright cups lie in order along one curve and tipped cups lie in order along another curve away from it.}
\label{fig:latent}
\end{figure*}

A world model predicts the outcome of actions so that an agent can plan before it acts.
Models that predict pixels spend much of their capacity on details that do not matter for control~\citep{ha2018world,hafner2020dreamer,hafner2023dreamerv3,bruce2024genie}.
Joint-embedding predictive architectures avoid this by predicting the next embedding instead of the next image~\citep{assran2023ijepa,assran2025vjepa2,hansen2024tdmpc2,leworldmodel2026}, and they have become the standard way to learn a world model without reconstruction.

Predicting embeddings, however, has a trivial solution.
If the encoder maps every observation to the same point, the prediction error is zero~\citep{grill2020byol,chen2021simsiam}.
Every latent world model therefore adds a mechanism against collapse, such as a slowly moving target network with a stop-gradient~\citep{grill2020byol,assran2023ijepa}, a penalty on variance and covariance~\citep{bardes2022vicreg}, or a penalty that pushes the embedding distribution toward a Gaussian~\citep{balestriero2025lejepa,leworldmodel2026}.
Each of them brings a weight or a schedule to tune, and none of them changes the fact that the encoder is trained by the prediction loss, which is where collapse comes from.

Preventing collapse does not give the distances a meaning, yet planners in latent world models choose actions by these distances, from the predicted embedding to the goal.
A cup that slides and a cup that tips over look similar, but the slide can be undone and the tip cannot, so a planner needs the two outcomes to lie far apart.
The existing mechanisms place them close together, and the planner cannot avoid irreversible events.
\cref{fig:cup} shows the consequence.
With each of the three mechanisms, planning fails more often as the goal moves away, and at forty steps it fails in 38--52\% of trials in Cup and 41--76\% in Cliff, mostly through irreversible events.
\cref{fig:latent} shows the cause, since each mechanism mixes tipped cups with upright ones in the embedding space.

We propose Sense, which solves both problems with one loss.
Its distance is directed, so the distance from $x$ to $y$ differs from the distance from $y$ to $x$.
A ReLU on the coordinate differences achieves this, because the distance from $x$ to $y$ then counts only the coordinates where $y$ is larger than $x$.
The exponential of the negative distance, normalized over a set of candidate states, is the probability that the agent reaches $y$ after $x$, and the encoder is trained with one cross-entropy whose target is a state reached later in the same trajectory.

This loss avoids collapse on its own, because an encoder that tells future states apart has a lower loss than one that gives every candidate the same probability.
The distance it learns measures reachability.
A state that is often reached from $x$ lies near $x$, and a state that is never reached from $x$ lies far away.
An irreversible transition is never reversed in the data, so its return distance becomes large without any label.
Since the encoder does not depend on the predictor, it is trained first, and the predictor is then fitted to the fixed embeddings by plain regression.

\INTRORESULT

\section{Related Work}
\label{sec:related}

\paragraph{World models.}
A world model predicts the consequences of actions so that an agent can plan or learn a policy inside it~\citep{ha2018world}.
The Dreamer family learns such a model by reconstructing observations~\citep{hafner2020dreamer,hafner2023dreamerv3}, and large video models extend reconstruction to open-ended interactive generation~\citep{bruce2024genie}, so these models represent every visible detail whether or not it matters for control.
MuZero and TD-MPC2 avoid reconstruction by predicting reward and value~\citep{schrittwieser2020muzero,hansen2024tdmpc2}, which ties the representation to a reward.
Joint-embedding predictive architectures predict the embedding of a future input from the embedding of its context~\citep{lecun2022path,assran2023ijepa}, and V-JEPA~2 and LeWorldModel plan in such an embedding with an action-conditioned predictor~\citep{assran2025vjepa2,leworldmodel2026}.
Their encoder is trained by the prediction loss together with a mechanism against collapse.
DINO-WM keeps a pre-trained DINOv2 encoder fixed and trains only the predictor~\citep{zhou2025dinowm,oquab2024dinov2}, but its features know nothing about the dynamics of the environment.
Sense needs no reward and also fits the predictor to fixed embeddings, but its encoder is trained on the trajectories of the environment.

\paragraph{Representation collapse.}
A joint-embedding objective that only pulls related inputs together is minimized by a constant encoder.
Contrastive methods avoid this by pushing unrelated inputs apart~\citep{oord2018cpc,chen2020simclr}, asymmetric methods by a target network or a stop-gradient~\citep{grill2020byol,chen2021simsiam}, and regularization methods by a penalty on batch statistics, such as the variance and covariance penalty of VICReg~\citep{bardes2022vicreg} and the Gaussian penalty SIGReg used by LeJEPA and LeWorldModel~\citep{balestriero2025lejepa,leworldmodel2026}.
In a latent world model these remedies are added to the prediction loss with a weight or a schedule, and since they rely on batch statistics rather than on the temporal order of states, their distances do not reflect the dynamics.
Sense is contrastive in form, but its positives are states reached later in the same trajectory, and this one loss is the whole objective of the encoder.

\paragraph{Temporal distances.}
Representations for control have long used the order of states.
Time-contrastive networks pull together frames that are close in time~\citep{sermanet2018tcn}, and VIP and HILP learn embeddings whose Euclidean distances approximate a goal-conditioned value~\citep{ma2023vip,park2024hilp}.
These distances are symmetric, whereas the time to reach a goal is not.
Contrastive reinforcement learning estimates the probability of reaching a goal with a critic over state, action and goal~\citep{eysenbach2022contrastive}, quasimetric models build the asymmetry into the architecture~\citep{wang2022quasimetric,wang2022iqe}, and QRL learns it by constraining each observed transition to unit length~\citep{wang2023qrl}.
All of them solve goal-reaching tasks with a value function or a policy trained beside a representation.
Sense uses a loss of the same family for a different purpose.
It is the only objective of the encoder of a world model, where it replaces the mechanism against collapse, and its quasimetric $\sum_i\relu(y_i-x_i)$ needs no network beyond the encoder.

\paragraph{Reversibility and safety.}
Irreversible events are a central concern of safe reinforcement learning~\citep{garcia2015safe}.
Reversibility has been estimated by learning a policy that resets the environment~\citep{eysenbach2018leave} and by a classifier on the temporal order of observations~\citep{grinsztajn2021reversibility}, and relative reachability penalizes actions that shrink the set of states the agent can still reach~\citep{krakovna2018side}.
Each of these methods trains a separate estimator beside the agent.
Sense trains none, because the return distance of a transition, measured in the embedding that the world model already uses, is its estimate of how hard the transition is to reverse.

\section{Method}
\label{sec:method}

\subsection{Preliminaries}

A latent world model has an encoder $f$, which maps an observation $o_t$ to an embedding $z_t=f(o_t)\in\mathbb{R}^D$, and a predictor $g$, which maps $(z_t,a_t)$ to an estimate of $z_{t+1}$.
If the encoder is trained by the prediction loss $\|g(z_t,a_t)-z_{t+1}\|^2$ alone, a constant encoder minimizes it, which is why the mechanisms of \cref{sec:related} are added.
Sense trains the encoder with a different loss, which we build in three steps, a directed distance, its normalization into a probability, and a cross-entropy.

\subsection{Directed distance}

Let $x,y\in\mathbb{R}^D$ be the embeddings of two states, where the arrow in $x{\to}y$ denotes the direction from $x$ to $y$.
We define the distance from $x$ to $y$ as
\begin{equation}
\dto{x}{y}=\sum_{i=1}^{D}\relu(y_i-x_i),
\label{eq:d}
\end{equation}
the cost of going from $x$ to $y$.
Only the coordinates in which $y$ is larger than $x$ contribute, so the cost $\dto{y}{x}$ of going back is in general different.
The distance is zero from a point to itself and satisfies the triangle inequality $\dto{x}{y}\le\dto{x}{w}+\dto{w}{y}$ because $\relu$ is subadditive, but it is not symmetric.

Writing $h(z)=\sum_i z_i$ for the height of an embedding,
\begin{equation}
\dto{x}{y}=\tfrac12\|y-x\|_1+\tfrac12\big(h(y)-h(x)\big).
\label{eq:height}
\end{equation}
The first term is an ordinary distance.
The second is a difference of heights, and it makes the two directions differ, since $\dto{y}{x}-\dto{x}{y}=h(x)-h(y)$.
The asymmetry is therefore set by one scalar per state, the sum of its coordinates, and needs no extra output of the network.

\begin{figure*}[t]
\centering
\includegraphics[width=0.8\textwidth]{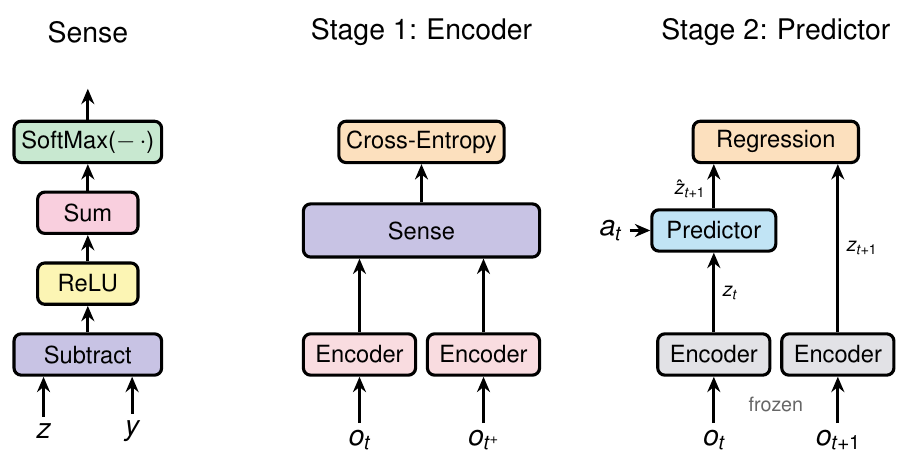}
\caption{Left: Sense. $z$ is the embedding of the current state and $y$ of a later one. Right: the two training stages of a world model built on Sense. Stage 1 trains the encoder with Sense alone. Stage 2 freezes the encoder and fits the predictor $g$ to its embeddings by regression.}
\label{fig:sense}
\end{figure*}

\subsection{Sense}

We define
\begin{equation}
\Sense(x{\to}y)=e^{-\dto{x}{y}},
\label{eq:sense}
\end{equation}
which measures how easily the agent goes from $x$ to $y$.
It lies in $(0,1]$ and equals $1$ only when $y$ can be reached from $x$ at no cost.
By the triangle inequality it multiplies along a path,
\begin{equation}
\Sense(x{\to}y)\;\ge\;\Sense(x{\to}w)\,\Sense(w{\to}y),
\end{equation}
so if each step $t$ of a path keeps a fraction $\gamma_t=e^{-c_t}$ of the sense, the whole path keeps $\prod_t\gamma_t=e^{-\sum_t c_t}$.
A discount factor is the special case in which every $\gamma_t$ is the same constant chosen by hand, whereas in Sense each step has its own factor, and the encoder learns it.

Given a set of candidate states $y_1,\dots,y_B$, Sense is normalized into a distribution,
\begin{equation}
p(y_j\mid x)=\frac{\Sense(x{\to}y_j)}{\sum_{k}\Sense(x{\to}y_k)},
\label{eq:p}
\end{equation}
the probability that $y_j$ is, among the candidates, the state the agent reaches after $x$.

\subsection{Training}

Let $o_t$ be an observation and $o_{t^+}$ an observation from a later step of the same trajectory, sampled uniformly from the remaining steps.
For a batch of $B$ such pairs with $z_b=f(o_{t_b})$ and $y_b=f(o_{t_b^+})$, the loss is
\begin{equation}
\mathcal{L}=-\frac1B\sum_{b=1}^{B}\log\frac{e^{-\dto{z_b}{y_b}}}{\sum_{k=1}^{B}e^{-\dto{z_b}{y_k}}}.
\label{eq:loss}
\end{equation}
This is the only loss of the encoder, and it takes three lines of code.
\begin{sensecode}
# embed a state and a later one
z, y = f(o_now), f(o_later)
# distance from every z to every y
d = relu(y[None] - z[:, None]).sum(-1)
# the state that was reached is the answer
loss = cross_entropy(-d, arange(B))
\end{sensecode}
It has no extra hyperparameter, since the encoder sets the scale of the embedding and the future is sampled uniformly.

Because the loss does not involve the predictor, training has two stages.
The encoder is trained first.
The predictor $g$ is then trained to map $z_t$ and $a_t$ to $z_{t+1}$ on the embeddings of the trained encoder, so its targets no longer change and its loss cannot change the encoder.
On TwoRoom, where we use the official code, the predictor is trained in the same run on embeddings whose gradient is blocked, which leaves the encoder unchanged as well.

\subsection{Why the embedding cannot collapse}

\begin{proposition}
\label{prop:collapse}
Let $m(f)=\frac1B\sum_{b}\big(\frac1B\sum_{k}\dto{z_b}{y_k}-\dto{z_b}{y_b}\big)$ be the margin of an encoder $f$ on a batch, the amount by which the reached state is closer than an average candidate.
If $f$ is constant, $\mathcal{L}=\log B$.
For any constant $a$ and any encoder $f$, the encoder $f_t=a+t f$ satisfies $\mathcal{L}(f_t)=\log B-t\,m(f)+O(t^2)$ as $t\to0^+$.
Hence, if any encoder has a positive margin, every neighborhood of a constant encoder contains an encoder with a lower loss, and no constant encoder is a local minimum of $\mathcal{L}$.
\end{proposition}
A constant encoder gives every candidate the same sense, so \cref{eq:p} is uniform and the loss equals $\log B$, the loss of a uniform guess.
Moving away from it toward any encoder that places reached states closer than average lowers the loss at first order.
Under a prediction loss the opposite holds.
There the constant encoder predicts perfectly and is the global minimum, which is why an extra mechanism is required.
Under Sense the loss descends away from the constant encoder, and no extra mechanism is needed.
The proposition concerns the loss and not the optimizer.
With the convention $\relu'(0)=0$ an exactly constant encoder receives no gradient, but a randomly initialized encoder is not constant, and no run in our experiments collapsed.

\subsection{What the distance means}

\Cref{eq:loss} is a contrastive loss~\citep{oord2018cpc}.
Let $p^+(y\mid x)$ be the probability that $y$ is reached at a later step of a trajectory through $x$, and $p(y)$ the probability of $y$ in the data.
If the score $-\dto{x}{y}$ could be replaced by an arbitrary function $s(x,y)$, the minimizer of the loss would be the log density ratio
\begin{equation}
s(x,y)=\log\frac{p^+(y\mid x)}{p(y)}+c(x),
\label{eq:opt}
\end{equation}
where $c(x)$ does not depend on $y$.
Sense cannot express an arbitrary score.
By \cref{eq:height}, every score it can express is $-\tfrac12\|y-x\|_1-\tfrac12\big(h(y)-h(x)\big)$, a symmetric part plus the difference of one scalar potential, so the encoder learns the member of this family that best fits \cref{eq:opt}, and the fit is exact only when the density ratio has this form.
Within this family the distance from $x$ to $y$ is small when $y$ is reached from $x$ more often than it is reached in general, and large when $y$ is rarely reached from $x$.

The restriction is the one that irreversibility needs.
By \cref{eq:height}, $\dto{z_{s'}}{z_s}-\dto{z_s}{z_{s'}}=h(z_s)-h(z_{s'})$, so a transition is harder to undo than to make exactly when it lowers the height.
Consider a transition $s\to s'$ that cannot be reversed.
In the data $s'$ follows $s$, but $s$ never follows $s'$.
The loss therefore treats $s'$ as a reached state for $s$ and pulls the forward distance $\dto{z_s}{z_{s'}}$ down, while $s$ appears only as a negative candidate for $s'$, which pushes the return distance $\dto{z_{s'}}{z_s}$ up.
For a reversible transition both directions appear in the data, and both distances are pulled down.
The encoder can meet both demands of an irreversible transition with the height alone, by placing $s'$ below $s$, and no label is needed to mark the transition.
We measure the danger of a transition by its return distance,
\begin{equation}
\mathrm{danger}(s\to s')=\dto{z_{s'}}{z_s},
\label{eq:danger}
\end{equation}
which a symmetric distance cannot represent, because it assigns the same length to both directions.

\subsection{Planning}

To reach a goal $z_g$, the agent predicts the final embedding $\hat z$ of each candidate action sequence and chooses the sequence whose $\hat z$ minimizes \PLANSCORE.
We add no safety term.
After an irreversible event the distance back to the goal is large, so the planner avoids such events by itself.

\subsection{Comparison with existing mechanisms}
\label{sec:why}

\begin{table*}[t]
\centering
\caption{What each way of training a latent world model needs beside the encoder and the predictor, and what its distances mean.}
\label{tab:why}
\begin{tabular}{lcccccc}
\toprule
 & Extra network & Loss terms & Weights to tune & Needs predictor & Distance means & Direction \\
\midrule
EMA target & target encoder & 1 & momentum & yes & nothing & no \\
VICReg & none & 3 & 2 & yes & nothing & no \\
LeWorldModel & none & 2 & 1 & yes & nothing & no \\
\textbf{Sense} & none & 1 & 0 & no & reachability & yes \\
\bottomrule
\end{tabular}
\end{table*}

\Cref{tab:why} compares Sense with the mechanisms it replaces in four respects.

\paragraph{One term.}
A target network changes the optimization so that the collapsed solution is not reached in practice, while VICReg and SIGReg add penalties whose minimum is far from collapse and balance them against prediction with a weight.
Sense has a single term that both prevents collapse and gives the distance its meaning.
The denominator of \cref{eq:p} pushes apart states that do not follow one another, the numerator pulls together states that do, and the normalization balances the two without a weight.

\paragraph{No predictor in the loop.}
In the other methods the encoder is trained by the prediction loss, so the encoder and the predictor must be trained together, and each learns against a target that the other keeps moving.
With Sense the encoder is trained without the predictor, and fitting the predictor to embeddings that no longer change is an ordinary regression problem with no risk of collapse.
The predictor can be replaced or retrained without touching the encoder, and a planning failure can be traced to its source, the embedding or the predictor, as we do in \cref{sec:far}.

\paragraph{Distances with a meaning.}
After training with a penalty on batch statistics, the distance between two embeddings reflects how different the two images look.
After training with Sense, it is fitted to the negative log of how often one state leads to the other, as \cref{eq:opt} shows, and this is the quantity a planner needs.

\paragraph{Direction.}
Many transitions are easier in one direction than in the other, and an object is easier to break than to repair.
A Euclidean distance is symmetric and cannot represent this.
The directed distance represents it through the height $h$ of \cref{eq:height}, and \cref{eq:danger} measures it.

\paragraph{A note on the brain.}
The hippocampus has been described as a predictive map, in which a place is represented by the states that tend to follow it~\citep{dayan1993sr,stachenfeld2017hippocampus}.
The embedding learned by Sense has the same property, since by \cref{eq:opt} the position of a state is fitted to the states that follow it.

\section{Experiments}
\label{sec:exp}

\subsection{Setup}

\begin{figure}[t]
\centering
\includegraphics[width=\columnwidth]{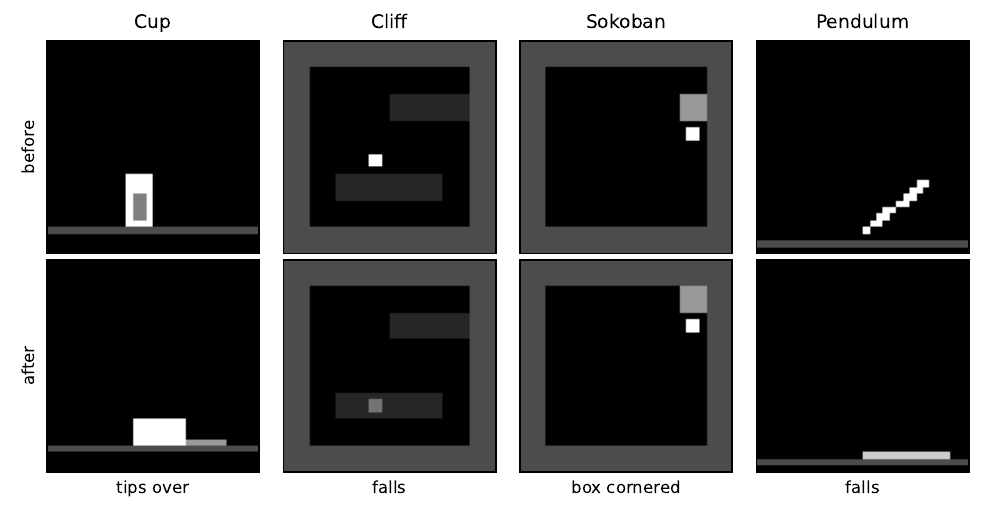}
\caption{The four environments. Each has one kind of irreversible transition.}
\label{fig:envs}
\end{figure}

\paragraph{Environments.}
We use five environments.
Four are rendered as $32{\times}32$ images, each with five actions and one kind of irreversible event (\cref{fig:envs}).
In Cup a hard push can tip a cup over, and in Cliff an agent on a grid can walk off an edge.
In Sokoban a box pushed against a wall or into a corner can no longer reach most of the grid, although nothing in the image signals this.
In Pendulum a pole that leans too far falls and stays down.
The data are 50{,}000 transitions of a uniformly random policy.
The fifth environment is TwoRoom from the LeWorldModel benchmark~\citep{leworldmodel2026}, where an agent observed through $224{\times}224$ images navigates from one room through a door to a goal in the other.
TwoRoom has no irreversible event and is used for planning.
Details are in \cref{app:envs}.

\paragraph{Compared methods.}
We compare Sense with an EMA target network, VICReg~\citep{bardes2022vicreg} and LeWorldModel~\citep{leworldmodel2026}, which trains the encoder and the predictor jointly with a prediction loss and SIGReg.
All four share the convolutional encoder, the predictor, the data, the number of updates and the evaluation, so only the training objective of the encoder differs.
The three baselines measure distance with the Euclidean norm, the distance their objectives are written in, and Sense uses \cref{eq:d}.
Every number is the mean over three seeds, and the tables also give the standard deviation.
On TwoRoom we take the official LeWorldModel code, train its encoder with Sense in place of the prediction loss and SIGReg, and evaluate every model with three evaluation seeds.

\paragraph{Measures.}
\emph{Danger AUC} ranks the transitions that change the state by \cref{eq:danger}, or by step length for the baselines, and reports the area under the ROC curve for separating irreversible transitions from the rest.
Labels are used only for this evaluation.
\emph{Planning} gives the agent a goal three steps away, searches all $125$ action sequences with the predictor, and counts a success when the agent reaches the goal without an irreversible event.
On TwoRoom, planning follows the LeWorldModel protocol, with CEM over a horizon of five, a goal 25 steps ahead in a dataset trajectory, and a budget of 50 steps in each of 50 episodes.
\emph{Effective rank}~\citep{roy2007erank} of the embedding measures collapse.

\subsection{Results}

\begin{table*}[t]
\centering
\caption{Danger AUC and planning success on five environments. On the first four, all methods share encoder, predictor, data, updates and evaluation, and each number is the mean and standard deviation over three seeds. TwoRoom has no irreversible event, and every model is evaluated with three evaluation seeds. \citet{leworldmodel2026} report $0.87$ for LeWorldModel.}
\label{tab:main}
\MAINTABLE
\end{table*}

\MAINTEXT

\subsection{Farther goals}
\label{sec:far}

The test above places the goal three steps away, and we now move it farther.
The goal is the state reached after $H$ random safe steps.
At every step the agent scores each three-step action sequence by summing the distances from its three predicted embeddings to the goal, using the round-trip distance for Sense and the Euclidean distance for the baselines, and it executes the first action of the best sequence before replanning.
A trial lasts at most $3H$ steps and fails if the goal is not reached or an irreversible event occurs.
\cref{fig:cup} shows the failure rate in Cup as $H$ grows to 40 steps, and in Pendulum $H$ goes up to 15 because its episodes are short.
With every baseline the failure rate rises with the distance in every environment.
At the farthest goals it is 38--52\% in Cup, 41--76\% in Cliff, 29--68\% in Sokoban and 49--65\% in Pendulum, and in Cup and Cliff most of these failures are irreversible events.
With Sense it is 0\%, 1\%, 4\% and 33\%, and in Cup and Cliff it does not grow with the distance.

These failures are not caused by prediction errors.
The dashed lines of \cref{fig:cup} show the failure rate when the prediction is exactly correct under the true physics, and in Cup, whose dynamics are stochastic, each action sequence is then simulated eight times and the scores are averaged.
The failure rates of LeWorldModel and VICReg barely change, staying at 42\% and 41\% in Cliff and at 54\% and 46\% in Cup, while Sense stays near zero.
The failures therefore come from the distance in the embedding.
The Euclidean distance between embeddings is informative only near the goal, and far from it the action that reduces this distance most often leads through an irreversible event.
The round-trip distance of Sense includes the cost of returning, so it stays informative at long range.

\subsection{Greedy navigation around a cliff}
\label{sec:cliff}

The planning tests above search three steps ahead, which is enough to detect most dangers in advance, so we also test the agent without search.
In Cliff we take every pair of start and goal whose shortest safe path is longer than the straight one, so that the direct way crosses the cliff.
The agent moves greedily by one-step lookahead, taking at each step the action whose predicted next embedding is nearest to the goal, and no method receives any label for the cliff.

\begin{table}[t]
\centering
\caption{Greedy navigation in Cliff on goals whose direct path crosses the cliff, mean and standard deviation over three seeds.}
\label{tab:cliff}
\CLIFFTABLE
\end{table}

\CLIFFTEXT

\subsection{Is the direction needed?}
\label{sec:sym}

By \cref{eq:height}, the round-trip score used in planning equals $\|y-x\|_1$.
One may therefore ask whether the direction matters, or whether the same future sampling and the same cross-entropy with a symmetric distance would work as well.
We train this variant by replacing \cref{eq:d} with $\|y-x\|_1$ and keeping everything else unchanged.

\begin{table}[t]
\centering
\caption{Sense against the same loss with the symmetric distance $\|y-x\|_1$, mean and standard deviation over three seeds. Irreversible: fraction of trials toward the farthest goal of \cref{sec:far} that end with an irreversible event.}
\label{tab:sym}
\SYMTABLE
\end{table}

\SYMTEXT

\section{Conclusion}

We presented Sense, a loss that trains the encoder of a latent world model with a single cross-entropy over a directed distance.
It replaces the target network, the variance penalty and the distribution penalty, has no weight to tune, and trains the encoder without the predictor.
The embedding it produces does not collapse, its distances measure how reachable one state is from another, and the return distance of a transition measures how hard the transition is to reverse.
On four pixel environments Sense identifies irreversible events without labels.
As the goal moves farther away its planning failure rate stays near zero in Cup, Cliff and Sokoban and is the lowest of all methods in Pendulum, and on TwoRoom it reaches every goal.

The experiments in this paper are small in scale, but Sense requires nothing beyond an encoder and trajectories.
We plan to use it in place of the regularizer in large latent world models trained on video and robot data.

\bibliography{refs}
\bibliographystyle{icml2026}

\newpage
\appendix
\onecolumn

\section{Environments}
\label{app:envs}
All environments render $32{\times}32$ images and have five actions: no-op and four directions.
Data are 50{,}000 transitions from a uniformly random policy. An episode is reset every 40 steps, and after an irreversible event it is reset with probability $0.15$ per step.

\paragraph{Cup.}
The state is the cup position $x\in[0,1]$ and whether the cup has tipped.
A light push moves the cup by $0.06$ and a hard push by $0.12$.
A hard push tips the cup with probability $0.8$, after which the cup lies on its side, water appears on the table and no action has any effect.

\paragraph{Cliff.}
The agent moves on a $6{\times}6$ grid surrounded by a wall.
Seven cells are cliff, four in one row and three in another.
Entering a cliff cell is a fall, after which the agent is drawn in a different shade and cannot move.

\paragraph{Sokoban.}
An agent and a box share a $6{\times}6$ grid surrounded by a wall, with $1{,}260$ states in total.
Walking into the box pushes it one cell if the cell behind it is free.
A box in the interior can still reach all 36 cells, a box against a wall only the 6 cells along that wall, and a box in a corner only its own cell.
A push that reduces this number is irreversible.

\paragraph{Pendulum.}
The pendulum follows $\ddot\theta=9.8\sin\theta+u$ with $\Delta t=0.1$ and $u\in\{0,\pm4,\pm10\}$.
When $|\theta|>1$ the pole falls to the ground and stays there.
The image has two channels, the pole at the current and at the previous time step, drawn with sub-pixel interpolation so that the image determines angle and velocity.

\paragraph{TwoRoom.}
TwoRoom is the navigation environment of the LeWorldModel benchmark~\citep{leworldmodel2026}.
Two rooms are separated by a wall with a single door.
The agent starts in one room and must reach a goal in the other room.
The data and the evaluation protocol are those of LeWorldModel.

\section{Implementation}
\label{app:impl}
The encoder is three convolutions with stride 2 followed by two linear layers, with output dimension $D=32$.
The predictor is a residual MLP with three hidden layers of width 512 whose input and output are normalized by the per-coordinate mean and standard deviation of the embedding.
The encoder is trained for 12{,}000 updates with Adam, learning rate $3{\times}10^{-4}$ and batch size 256.
The predictor is then trained for 60{,}000 updates on the frozen encoder with learning rate $10^{-3}$ and a cosine schedule.
This schedule is the same for all methods.
For LeWorldModel the SIGReg weight is $0.1$ with 1{,}024 random projections, for VICReg the variance and covariance weights are $25$ and $1$, and for the target network the momentum is $0.99$.

On TwoRoom we use the official LeWorldModel code with the encoder trained by Sense and the predictor trained on embeddings whose gradient is blocked, train every model for ten epochs, and compute the batch-normalization statistics of Sense with its final weights.

\paragraph{Proof of \cref{prop:collapse}.}
The distance of \cref{eq:d} does not change when the same constant is added to both arguments, and it is positively homogeneous, so $\dto{a+tx}{a+ty}=t\,\dto{x}{y}$ for $t\ge0$.
Write $d_{bk}=\dto{z_b}{y_k}$ for the encoder $f$.
For $f_t$ the loss is
\[
\mathcal{L}(f_t)=\frac1B\sum_{b}\Big(t\,d_{bb}+\log\sum_{k}e^{-t\,d_{bk}}\Big),
\qquad
\log\sum_{k}e^{-t\,d_{bk}}=\log B-\frac tB\sum_k d_{bk}+O(t^2).
\]
Summing over $b$ gives $\mathcal{L}(f_t)=\log B-t\,m(f)+O(t^2)$, and $t=0$ gives the constant case.
If $m(f)>0$, then $\mathcal{L}(f_t)<\log B$ for all small $t>0$, and $f_t\to a$ uniformly as $t\to0$, so every neighborhood of the constant encoder $a$ contains an encoder with a lower loss.

\end{document}